\documentclass[runningheads]{llncs}
\usepackage[T1]{fontenc}
\makeatletter
\def\@chapapp{Chapter}
\makeatother

\usepackage{graphicx}
\usepackage[numbers]{natbib}
\usepackage{multicol}
\usepackage[bookmarks=true]{hyperref}

\usepackage{graphics} 
\usepackage{multirow}
\usepackage{makecell}
\usepackage{booktabs}
\usepackage{threeparttable}
\usepackage{diagbox}  
\usepackage{slashbox} 
\usepackage{amsmath}
\usepackage{algorithm} 
\usepackage{algpseudocode}
\usepackage{caption}
\usepackage{xcolor}
\usepackage{appendix}
\usepackage{comment}
\usepackage{siunitx}
\usepackage{float}
\usepackage{appendix}

\begin{document}
\title{One Fling to Goal: Environment-aware Dynamics for Goal-conditioned Fabric Flinging}

\titlerunning{One Fling to Goal: Goal-conditioned Fabric Flinging}
%
\author{Linhan Yang\inst{1,2,3} \and
Lei Yang\inst{1,2} \and
Haoran Sun\inst{1,3} \and
Zeqing Zhang\inst{1,2} \and
Haibin He\inst{2} \and
Fang Wan\inst{3} \and
Chaoyang Song\inst{3,*} \and
Jia Pan\inst{1,2,*}}
\authorrunning{Yang et al.}
%
\institute{The University of Hong Kong, Hong Kong SAR \and
Centre for Transformative Garment Production, Hong Kong SAR \and
Southern University of Science and Technology, Shenzhen, China}

\maketitle              

\begin{abstract}

    Fabric manipulation dynamically is commonly seen in manufacturing and domestic settings. While dynamically manipulating a fabric piece to reach a target state is highly efficient, this task presents considerable challenges due to the varying properties of different fabrics, complex dynamics when interacting with environments, and meeting required goal conditions. To address these challenges, we present \textit{One Fling to Goal}, an algorithm capable of handling fabric pieces with diverse shapes and physical properties across various scenarios. Our method learns a graph-based dynamics model equipped with environmental awareness. With this dynamics model, we devise a real-time controller to enable high-speed fabric manipulation in one attempt, requiring less than 3 seconds to finish the goal-conditioned task. We experimentally validate our method on a goal-conditioned manipulation task in five diverse scenarios. Our method significantly improves this goal-conditioned task, achieving an average error of 13.2mm in complex scenarios. Our method can be seamlessly transferred to real-world robotic systems and generalized to unseen scenarios in a zero-shot manner.

\keywords{Machine Learning in Robotics \and Manipulation \and Deformable Objects}
\end{abstract}

\section{Introduction}
\label{sec:intro}

\begin{figure}[!thb]
        \centering
        \includegraphics[width=0.9\textwidth]{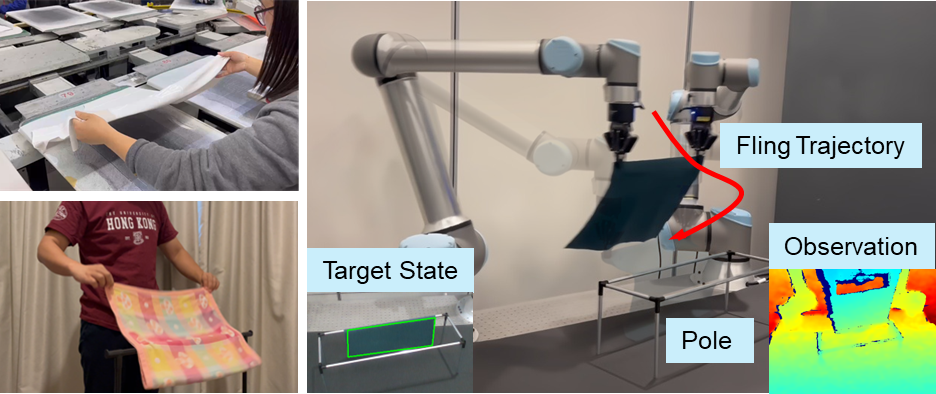}
        \caption{Goal-conditioned dynamic manipulation task in complex environments. (Left) Fabric manipulation in manufacturing and domestic settings. (Right) Experiment setup. We extract the depth information from an RGBD camera and adjust the picker movement in real-time to achieve the final target state. We also validate this policy across complex environments.
        }
        \vspace{-4mm}
        \label{fig:overview}
    \end{figure}

    Despite the underactuated nature of fabrics, humans can efficiently place a fabric piece in a specified state by leveraging the dynamics of fabrics. Numerous examples can be found in domestic settings, such as flinging cloth onto a drying pole or in industrial settings where an operator flings a fabric piece onto the board of a screen printing machine, as shown in Fig.~\ref{fig:overview}. While being a simple task for humans, it has been challenging for robots. Two critical problems are identified: 1) How can a robot leverage the fabric dynamics to achieve this task in a single shot like humans, and 2) How can a robot place the flung fabric object precisely in a specified state in different environments?

    Seminal works~\cite{berenson2013manipulation,mcconachie2020manipulating} proposed a model-free method to manipulate deformable objects with the assumption that the grippers move quasi-statically. 
    Unlike quasi-static fabric manipulation~\cite{yan2021learning, zhang2022learning, hoque2020visuospatial, shen2023action} that gently moves a fabric object, dynamic manipulation of fabrics is investigated in this study as it makes use of acceleration to improve robots' action efficiency and physical reachability, allowing robots to achieve a specified target state of the fabric object with as few as a single interaction~\cite{mason1993dynamic}. However, exploiting the dynamics would come at the price of modeling complex, non-linear deformation of the fabrics under this high-speed manipulation.

    In recent years, many approaches have been proposed for dynamic fabric manipulation, ranging from cloth unfolding~\cite{ha2022flingbot, chen2022efficiently}, goal-conditioned flinging~\cite{chi2022iterative}, to multi-step manipulation for folding~\cite{canberk2023cloth}. Inspired by these related works and human dynamics when flinging fabric, our method generates a fling-and-then-pull trajectory, where the fling and pull actions transition at a critical turning point. Unlike FlingBot~\cite{ha2022flingbot}, which manually defines the turning point, our approach robustly estimates it to generalize to specified goal conditions.
    
    To achieve goal-conditioned tasks, one might choose the trial-and-error approach proposed in Iterative Residual Policy (IRP)~\cite{chi2022iterative}, which iteratively minimizes the residual between the target and the trial to find a desirable flinging trajectory. Alternatively, we propose using an online controller that minimizes the error between the predicted roll-out result and the goal state. This allows our method to accomplish goal-conditioned manipulation tasks with \textit{only one fling}, rendering the multiple trials in IRP unnecessary.
    
    While previous works~\cite{ha2022flingbot,chen2022efficiently,chi2022iterative,canberk2023cloth} focus on placing a fabric piece on a flat plane, goal-conditioned manipulation tasks in daily life are diverse, such as hanging a fabric on a drying pole or covering a table with a cloth. Therefore, our method is designed to be environmentally aware, enabling it to cope with various environments, as depicted in the bottom left of Fig.~\ref{fig:overview}.

    This paper presents an environment-aware approach to enable robots to achieve a goal-conditioned task in various environmental setups through a single dynamic fling. By \textit{goal-conditioned}, we refer to aligning the final state of the manipulated fabric with a pre-specified target state. The goal state of a fabric can be specified on a pole or an elevated platform, in addition to a simple flat plane, as shown in Fig.~\ref{fig:overview}. We term this property as \textit{environmental awareness}. Our approach achieves the goal state of the fabric with a \textit{single} attempt instead of combining multiple actions or conducting several trials. We use a model-based method to optimize the fling trajectory parameterized by the turning point to achieve this task. Since it is non-trivial to model deformable objects, especially in complex environments, analytically, we use a data-driven method to train a dynamics model with environmental information embedded. We also developed a real-time control policy to handle unknown fabric properties.

    \noindent \textbf{Dynamics model with environmental awareness. } 
    First, our approach includes a dynamics model that can accurately simulate the high-speed motion of fabrics with different physical properties. We achieve this by training the dynamics model to predict the velocity of a graph-based representation of the fabric at the next timestamp. To make our dynamics model environmentally aware, we incorporate the signed distance and its gradient at each node of the graph-based representation. This allows the learned dynamics model to accurately predict the velocity at each node, even in environmental setups such as a pole or a tabletop. It is important to note that our dynamics model is trained without specific goal conditions, making it versatile and applicable to various object properties and environmental settings.

    \noindent \textbf{Model-based real-time control policy.} 
    A control policy is necessary to place a fabric object with unknown physical properties into a specified final state with a single attempt. Due to fabrics' high degrees of freedom and the large range of possible actions, we employ a fling-and-pull strategy frequently observed in human demonstrations. We propose a two-stage approach that achieves a real-time control policy during execution, ensuring that a single attempt is sufficient to reach the goal state. First, using the learned dynamics model, we \textit{virtually} sample several action trajectories with different turning points. Among these samples, we select the trajectory with the final state closest to the goal as the initial guess for execution. However, since the dynamics model is unaware of the physical properties of the fabric, the initial trajectory is less likely to place the fabric precisely in the specified goal state. Therefore, during execution, we employ a model predictive control (MPC) method to adjust the fling trajectory in real time. This controller takes the current visual observation of the fabric as input and adjusts the velocities of both end-effectors (referred to as pickers) in the next time step. This real-time refinement enables the robot to achieve the goal-conditioned dynamic fling with just one attempt.

    We validate our approach in five scenarios, each with a different rigid object as its environmental setup: a flat plane, an elevated platform, a hemisphere, a pole, and an unseen tabletop. Our results demonstrate that the proposed method outperforms other state-of-the-art approaches using the fling action for dynamic fabric manipulation. We conduct ablated studies to validate our design choices, such as environmental awareness and the real-time control policy. The experimental results show that our method can be generalized to a held-out scenario (the tabletop), fabrics with different physical properties, and varied environmental setups. Additionally, we conduct real-world experiments in a zero-shot manner. Our contributions can be summarized as follows:
    \begin{enumerate}
        \item The first approach to achieve goal-conditioned dynamic fabric manipulation in a single attempt.
        \item A dynamics model with environmental awareness to accurately predict fabric motion and environmental interaction.
        \item A model predictive control framework to adjust the fling trajectory in real-time, enabling precise dynamic manipulation.
    \end{enumerate}

\section{Related Works}
\label{sec:related works}

    This section discusses related studies on goal-conditioned manipulation, dynamic manipulation of deformable objects, and graph-based representation for fabric manipulation.

    \noindent\textbf{Goal-conditioned Manipulation. }
    Goal-conditioned manipulation tasks involve changing the state of objects to achieve a predefined goal state. Deformable objects present particular challenges due to their complex dynamics and a high degree of freedom~\cite{yin2021modeling}. Prior research has made notable advancements in deformable object manipulation tasks, such as water scooping \cite{niu2023goats}, rope manipulation~\cite{sundaresan2020learning, yin2021modeling}, fabric folding~\cite{weng2022fabricflownet}, and unfolding~\cite{ha2022flingbot, zhang2022learning,chen2022efficiently}. However, these methods primarily focus on manipulation within a flat plane, simplifying the interaction between the object and its environment and confining manipulation to a two-dimensional plane. Our research aims to extend these methodologies by introducing a closed-loop fling policy that operates in more complex environments where the goal state could involve rich interaction with curved surfaces.

    \noindent\textbf{Dynamic Manipulation. }
    Dynamic manipulation leverages high-speed actions such as tossing~\cite{zeng2020tossingbot}, swinging~\cite{wang2020swingbot}, and blowing~\cite{xu2022dextairity} to enhance the efficiency and extend the workspace of robotic systems, contrasting with quasi-static mechanisms that rely on sequential pick-and-place actions. Previous research in this domain has primarily focused on predicting picking positions before flinging actions~\cite{ha2022flingbot} or required multiple trials to adjust trajectories based on delta dynamics~\cite{lin2022learning}. Our work aims to develop a learning-based control policy that optimizes trajectory in real-time, reflecting a significant step forward in the dynamic manipulation of deformable objects.

    \noindent\textbf{Graph-based Representation for Fabric Manipulation. }
    Representing a fabric as a graph has emerged as a popular method in various robotics domains, including manipulation, locomotion, and physical modeling. Works such as~\cite{wang2018nervenet, yang2023tacgnn, ghasemipour2022blocks, sun2023bridging, zhang2024cafknet} demonstrate the effectiveness of this approach. The particle-based nature of graph representations offers significant advantages in modeling cloth dynamics, leveraging the inductive biases of particle systems to capture the behavior of deformable objects more accurately. Additionally, this representation is invariant to visual features, facilitating sim-to-real transfer. Contrastingly, previous studies have either relied on image-based representations that predict future states of cloth after an action~\cite{zeng2021transporter, shen2023action, seita2021learning} or utilized fixed-size latent vectors for state representation~\cite{zhou2021lasesom}. These methods often struggle to generalize across fabrics of varying shapes and properties.
        
    \begin{figure}[!th]
        \centering
        \includegraphics[width=0.9\textwidth]{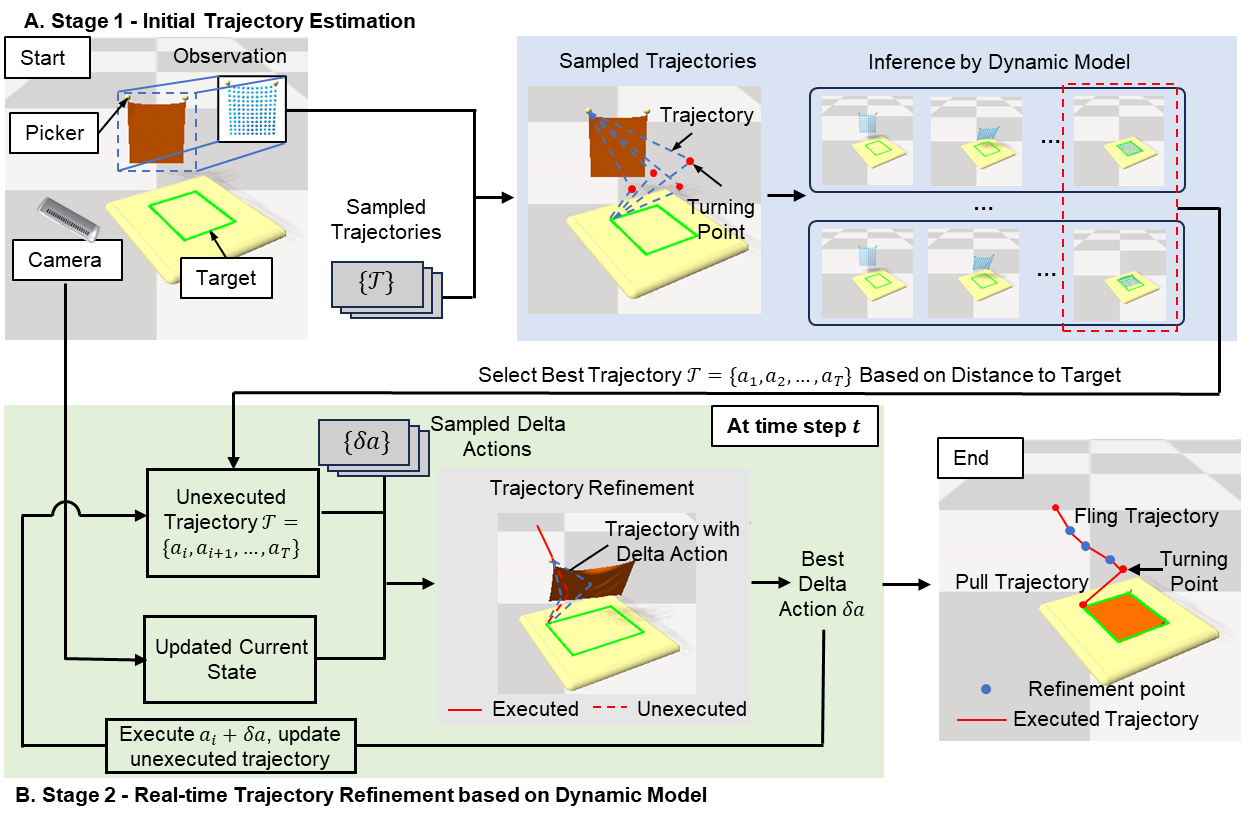}
        \caption{Overview of our proposed algorithm. The first stage (top) involves sampling a batch of picker trajectories based on the current and target states before executing the action. The dynamics model predicts the cloth state after execution, and the distance to the target state is evaluated using Eq.~\ref{eq:distance}. The best trajectory is selected. In the second stage (bottom), fine adjustments are made at each action based on the current visual observation, allowing for either more forceful or gentler movements.
        }
        \label{fig:method}
        \vspace{-8mm}
    \end{figure}
    
\section{Methodology}

    \subsection{Problem Formulation}
    \label{sec:problem statement}
    
    \noindent \textbf{Overview.} \label{sec:3.1}
    Given a fabric piece with an unknown physical property stably grasped by two pickers at its corners (as shown in Fig.~\ref{fig:overview}), this study proposes a dynamic manipulation method to transform the given fabric into a user-specified goal state of the fabric with arbitrary position and orientation. We propose a learned dynamics model with environmental awareness to handle fabric pieces with varying physical properties, such as elasticity, and different environmental setups where rigid objects, such as poles or elevated platforms, have arbitrary positions and orientations in the robot base frame. Furthermore, we present a real-time model predictive control algorithm based on the learned dynamics model with environmental awareness to achieve goal-conditioned dynamic fabric manipulation tasks.
        
    \noindent \textbf{State representation.}
    We use a graph-based representation proposed in~\cite{lin2022learning} to represent the fabric's state, which is robust for sim2real transfer. The fabric's state at time $t$ is denoted $S_t = \langle V_t, E_t \rangle$, where $V_t=\{\mathbf{v}_i^t\}_{i\in[N_t]}$. For each node $i$, in addition to its spatial coordinate $\mathbf{x}_i$, a feature vector $\mathbf{p}_i$ is associated, containing history velocities and environmental information. The distance between the current state $S$ and the goal state $S_g$ is defined as the averaged distance between corresponding points $\mathbf{x}_i$ and $\mathbf{x}_i^g$ derived from bipartite graph matching, as in~\cite{lin2022learning}. Specifically, the distance is calculated as:
    \begin{equation}
        \label{eq:distance}
        \mathcal{D}(S,S_g) = \frac{1}{N} \sum_i \|\mathbf{x}_i - \mathbf{x}_i^g\|_2, \quad i\in[N].
    \end{equation}
    To achieve the goal-conditioned fabric manipulation task, we minimize the following cost function:
    \begin{equation}\label{eq:cost}
    \mathcal{L} = \min_{\mathcal{T}} \mathcal{D}(S_T, S_g),
    \end{equation}
    to find a trajectory $\mathcal{T}$ that minimizes the deviation between its final state $S_T$ and the specified goal $S_g$.

    \noindent \textbf{Action space. }
    We assume that the cloth is gripped by two pickers that move in a synchronized manner and consider a three-dimensional movement for two pickers. Intuitively, the action at each time step can be naively defined as the delta movement of both pickers:
    \begin{equation}
        \label{eq:action}
        \mathbf{a} = (\Delta x_l, \Delta y_l, \Delta z_l, \Delta x_r, \Delta y_r, \Delta z_r),
    \end{equation}
    where subscripts $l,r$ denote the left picker and right picker, respectively. The trajectory is a sequence of actions $\mathcal{T} = \{\mathbf{a}_1, \mathbf{a}_2,...,\mathbf{a}_T\}$. However, this naive action parameterization is high-dimensional, lacking natural constraints for valid actions. For instance, it does not consider that the left and right pickers should maintain a constant distance $L$ and move in sync to prevent the fabric from being under or over-stretched, which is undesirable for high-quality fling.

    To respect these constraints and reduce the search space of actions, we reformulate the action $\mathbf{a}$ as the motion of the virtual link connecting the left and right pickers. This motion involves the translation of the link's midpoint and the link's angular change around the $z$-axis of the robot's base frame. Assuming that the pickers move in sync, the angular velocity is constant. It can be calculated as $\Delta \theta = \frac{\theta_{g} - \theta_{s}}{T}$, where $\theta_{g}$ and $\theta_{s}$ represent the goal and initial angular offset, respectively, and $Z$ is the number of time steps. The control variable is the translational velocity, or delta movement, of the midpoint $\mathbf{m}$ from its initial position to its target. We denote this delta movement as $(\Delta m_x, \Delta m_y, \Delta m_z)$. Therefore, the action can be simplified to the midpoint movement, $\mathbf{a} = (\Delta m_x, \Delta m_y, \Delta m_z)$. The trajectory of the pickers can be easily derived from the trajectory of the midpoint, considering the constants $L$ and $\Delta \theta$. Moving forward, we will focus exclusively on the midpoint trajectory.

        
    To simplify the fling trajectory parameterization, we follow previous work~\cite{chi2022iterative} and use via-points. These include the start point $\mathbf{m}_s$, the turning point $\mathbf{m}_m$ in the middle, and the target point $\mathbf{m}_g$. The turning point is crucial as it marks the transition from flinging forward to pulling backward. The pickers fling the fabric forward to the turning point, accelerating at \SI{2}{m/s^2} and then decelerating at \SI{-2}{m/s^2}. They reach a static state at the turning point. Similarly, during the backward pull, the pickers accelerate and then decelerate at the same rate. Therefore, the trajectory can be easily derived once the turning point is determined.
        
    Next, we define a manipulation plane $\Pi$ spanned by $\overline{\mathbf{m}_s\mathbf{m}_g}$ and the $z$-axis of the robot base frame. If the turning point moves on either side of the plane $\Pi$, such action will only shear the fabric, leading to undesired folds. Thus, we constrain the midpoint to move in plane $\Pi$, and thus the turning point $\mathbf{m}_m$ can be parameterized by a 2D point $\mathbf{u}_m \in \Pi$. In this way, the midpoint trajectory can be represented as $\mathcal{T}=f(\mathbf{u})$ with $\mathbf{u} \in \Pi$, leading to a much lower-dimensional search space for high-quality manipulation trajectories. 
    
    
        
    \begin{figure}[!th]
        \centering
        \includegraphics[width=0.9\textwidth]{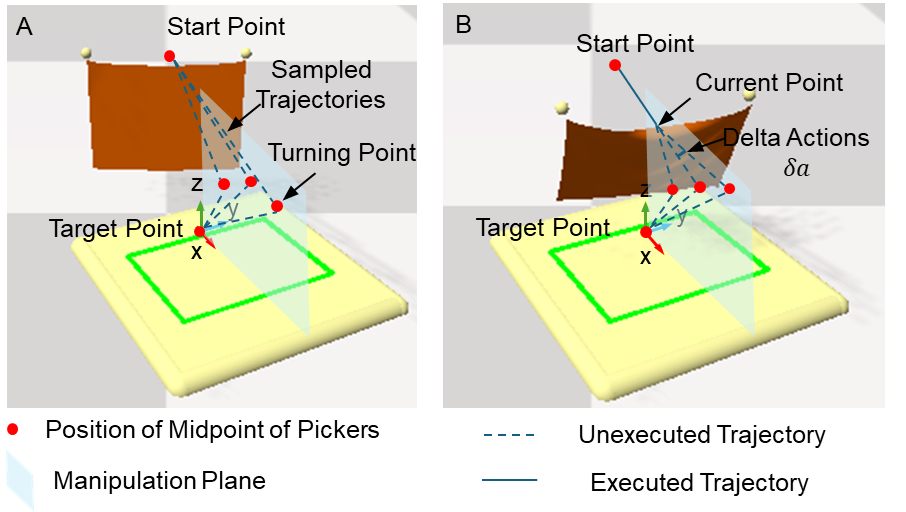}
        \caption{Reduced action space parameterized by the turning point. (A) In the first stage, a batch of trajectories parameterized by a turning point are sampled. (B) In the second stage, we refine the trajectory by adding delta actions and select the best one based on real-time visual feedback and a trained, dynamic model. 
        }
        \label{fig:method2}
    \end{figure}


    \subsection{Dynamics Model with Environmental Awareness}\label{sec:dynamics}

    A dynamic model is a crucial component of the model-based approach. Given the complexity of deformable object dynamics, we employ a data-driven method to train a generic dynamics model in a simulation environment using an autoregressive approach without requiring a specific goal condition. The dynamics model is designed to predict the next-timestep state across various environmental setups (to be introduced shortly).
    
    We use a graph neural network $\mathcal{G}_{\text{dyn}}$ with the state representation $S=\langle V, E \rangle$ to model the dynamics. Instead of directly predicting the next-timestep state of the fabric, the model estimates the velocity $\dot{\mathbf{x}}_i^t$ at each node $i$ at time $t$. The Euler method is then used to obtain the updated position $\mathbf{x}^{t+1}$:
    \begin{align}
        \label{eq:dynamics}
        \dot{\mathbf{x}}^{t+1} &= \mathcal{G}_{\text{dyn}}(S^t, \mathbf{a}_t | \Theta), \\ \nonumber
        \mathbf x^{t+1} &= \mathbf x^{t} + \dot{\mathbf{x}}^{t+1} \Delta t, 
    \end{align}
    where $\mathbf{a}_t$ is the action taken at the current time step and $\Theta$ denotes the set of trainable parameters of $\mathcal{G}_{\text{dyn}}$.

    \noindent\textbf{Feature vectors at graph nodes. }
    The feature vector at node $i$ (denoted $\mathbf{p}_i$) consists of two parts, i.e., the motion history and the environmental descriptor.
    It is worth noting that the physical properties of fabric are typically inaccessible. To accurately predict the velocity $\dot{\mathbf{x}}_i^t$ at node $i$, we store the velocities at $i$ from the previous $n$ time steps, represented by $\dot{X}_n = (\dot{\mathbf{x}}_i^t, \dot{\mathbf{x}}_i^{t-1}, \cdot\cdot\cdot, \dot{\mathbf{x}}_i^{t-n})$. Therefore, each node $i$ in the graph-based state $S$ stores its velocity history $\dot{X}_n$, part of the introduced feature $\mathbf{p}_i$. In this paper, $n$ is set to 5.
        
\begin{figure}[!h]
        \centering
        \includegraphics[width=0.9\textwidth]{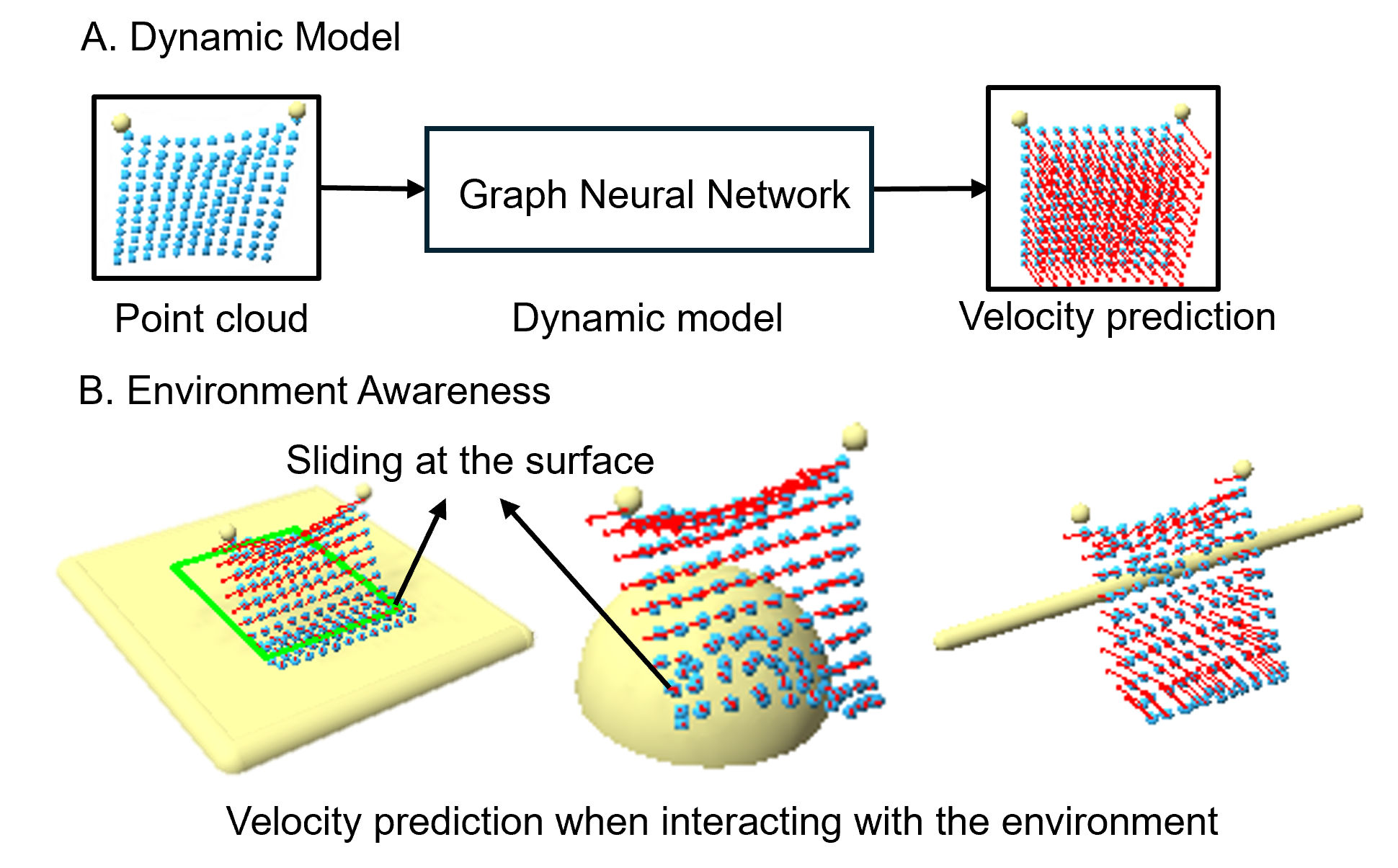}
        \caption{Dynamics model with environmental awareness. A) The state of the fabric is represented as a graph representation augmented with environmental information and is fed into the dynamics model. The velocity for each point in the graph is predicted. B) The dynamics model can predict the velocity of each point well by augmenting the fabric state with environmental information. Points in contact with the rigid objects in the environment will slide on the surface of the rigid object.}
        \label{fig:dynamic}
        \vspace{-8mm}
    \end{figure}

    The environment descriptor $\mathbf{e}_i$ is the other part of the feature $\mathbf{p}_i$ and enables the dynamics model to be aware of the future transition caused by the environment. For instance, our dynamics model should predict zero velocity for a node that collides with the surface of a rigid object in the environment at the current time step, as shown in Fig.~\ref{fig:dynamic}(B). The environment descriptor is defined as:
    \begin{equation}
        \mathbf{e}_i = (sdf(\mathbf{x}_i), sdf^\prime(\mathbf{x}_i))
    \end{equation}
    where $sdf(\mathbf{x})$ is the signed distance value at $\mathbf{x}$ and $sdf^{\prime}$ is the gradient of the signed distance field at this point\footnote{Gradients of a signed distance field at differentiable positions are unit-length vectors.}.
    \begin{align}
        sdf(\mathbf{x}_i) = \|\mathbf x_i-\mathbf q_j\|, \ \
        sdf^\prime(\mathbf{x}_i) = \frac{\mathbf x_i-\mathbf q_j}{\|\mathbf x_i-\mathbf q_j\|}, 
    \end{align}
    where $\mathbf q_j$ denotes the nearest point in the environment, such as a table or pole surface point, to the fabric graph node $\mathbf x_i$. Thus, an environment descriptor indicates the distance to the surface of a rigid object in the scene and the gradient direction pointing towards the scene.
    
    We concatenate $(\dot{X}_i, \mathbf{e}_i)$ to form the feature $\mathbf{p}_i$ at node $i$. With this fabric state representation as input, the proposed dynamics model is trained to predict the future motion of the fabric as well as complex interactions with the environment. At runtime, spatial coordinate $\mathbf{x}_i$ and nodal feature $\mathbf{p}_i$ are updated at each time step. 
    
    \noindent \textbf{Edge connectivity. }
    Finally, we define the graph edges that connect each node to other nodes in its vicinity as $E = \{ \|\mathbf{x}_i^t - \mathbf{x}_j^t\|_2 < r \}$, where $r$ denotes the distance threshold, and $\mathbf{x}_i^t$ and $\mathbf{x}_j^t$ refer to the positions of node $i$ and $j$ at time $t$. As the fabric moves in the air, the edge connectivity is updated at each time step to account for the evolving shape of the fabric.

     \noindent \textbf{Training loss.} We train the dynamics model to predict the velocity of the next time step by minimizing the following loss with stochastic gradient descent:
    \begin{equation}
        L = \min_{\Theta} \sum \| \mathcal{G}_{\text{dyn}}(S^t, \mathbf{a}_t|\Theta) - \dot{\mathbf{x}}^{t+1}_{GT} \|.
    \end{equation}
    
    \subsection{Controller for Real-time Trajectory Refinement}
    
    We observe human demonstrations and identify two major motion primitives: the flinging forward primitive and the pulling backward primitive. We mimic the human performance of this task by simplifying any potential trajectory to a \textit{fling-then-pull} movement, parameterized by a single turning point, as discussed in Sec.~\ref{sec:3.1}. However, this reduction alone is insufficient for achieving satisfactory results, as our learned dynamics model is assumed to be agnostic to the physical properties of the fabric. To accomplish the task, it is necessary to control the trajectory instantaneously based on real-time visual feedback. We develop a two-stage strategy to enable real-time control of the flinging trajectory using visual feedback, as shown in Fig.~\ref{fig:method2}.

    \noindent \textbf{Initial trajectory estimation.}
    In the first stage, we sample a coarse trajectory as an initial guess before action execution. The pickers' trajectory is generated by randomly sampling the turning point $\mathcal{T}=f(\mathbf{u} \in \Pi)$. For each trajectory, we infer the final state of the fabric using the learned dynamics model $\mathcal{G}_{\text{dyn}}$ in Sec.~\ref{sec:dynamics}. The roll-out trajectory whose final state minimizes Eq.~\ref{eq:objective_1} is selected as the initial estimation, as shown in Fig.~\ref{fig:method}A. In this stage, the control objective is:
    \begin{equation}
        \label{eq:objective_1}
        \min_{\mathcal{T}=f(\mathbf{u} \in \Pi)} \mathcal{D}\left(\mathcal{G}_{\text{dyn}}(S_s,\mathcal{T}), S_g \right),
    \end{equation}
    where $S_s, S_g$ denote the start state and goal state, respectively, and $x_t,z_t$ denote the coordinates of the turning point. Trajectory $\mathcal{T}$ can be derived from the turning point. Distance function $\mathcal{D}$ (Eq.~\ref{eq:distance}) calculates the distance between the final state of the trajectory $\mathcal{T}$ and the goal state $S_g$. To find the initial guess, we sampled 50 turning points and, thus, 50 trajectories in this stage.

    \noindent \textbf{Real-time trajectory refinement.}
    In the second stage, we perform fine-grained adjustments every 10 timesteps based on current visual observation through a model predictive controller while executing the initial trajectory. The control frequency was chosen as a trade-off between the inference cost and the accuracy. For each control time step, we sample $9$ delta actions, which are categorized into three types based on their impact on the trajectory:
    \begin{enumerate}
        \item Accelerate along the $x$- or $z$-axis of manipulation plane;
        \item Remain the same velocity;
        \item Decelerate along the $x$- or $z$-axis of manipulation plane.
    \end{enumerate}

    This results in a total of \(3 \times 3 = 9\) possible delta actions. Each delta action is added to the unexecuted trajectory, resulting in an updated turning point, as shown in Fig.~\ref{fig:method2}B. The updated trajectories are fed to the dynamics model to infer the final state of the fabric. The trajectory whose final state has a minimal distance from the target state is selected. 

    In this stage, the control objective is:
    \begin{equation}
        \label{eq:objective}
        \min_{\delta\mathbf{a}} \mathcal{D}(\mathcal{G}_{\text{dyn}}(S_c,\mathcal{T}_{u}^{\prime}=f(\mathcal{T}_u, \delta\mathbf{a})), S_g),
    \end{equation}
    where $\delta\mathbf{a}$ denotes the delta action added to the unexecuted trajectory $\mathcal{T}_u$, resulting in an updated trajectory $\mathcal{T}_{u}^{\prime}$. $S_c, S_g$ denote the current state and goal state, respectively. This two-stage control policy, leveraging our robust dynamics model, significantly improves the system's ability to adapt to novel environments and fabric with unknown properties.
    

\section{Experimental Results}
\label{sec:experiment}

    \subsection{Simulation Setup}
    The Nvidia Flex simulator, wrapped in SoftGym~\cite{lin2021softgym}, is utilized for data collection. Two pickers mimic the end-effectors of robot arms in the simulation and grasp a corner node of the cloth mesh. Trajectories of the pickers are generated, as described in the previous section, based on randomly sampled turning points. The pickers flip the fabric forward as it moves towards the turning point, then pull it backward to reach the target configuration. To accomplish this, the pickers should accelerate during the fling action, decelerating to zero at the turning point. During the pull action, the pickers should accelerate and then decelerate to reach the final target position. The physical properties of the fabric, such as density, elasticity, and size, are randomly sampled to enable our model to generalize to fabrics with unknown physical properties and minimize the sim-to-real gap.

    We considered five experimental scenarios in simulation:
    \begin{enumerate}
        \item \textbf{Flat Scenario}, where the objective is to place the cloth on a smooth, flat surface with a desired configuration. The absence of obstacles or complex terrain simplifies the cloth dynamics.
        \item \textbf{Platform Scenario,} which introduces a raised platform with a side length of \SI{0.4}{m}. The goal is to place the cloth at the center of the platform.
        \item \textbf{Hemisphere Scenario,} where a hemispherical object in a flat environment is introduced, similar to Platform Scenario. The objective is to cover the cloth on the hemisphere. The cloth may contact the sphere's curved surface, introducing complex interactions.
        \item \textbf{Pole Scenario,} where a long pole-like object is placed with a height of \SI{0.15}{m} and a width of \SI{0.4}{m}. The target is to place the cloth on the pole. The cloth may wrap around or interact with the pole during manipulation.
        \item \textbf{Table Scenario,} where the tabletop is placed with a height of \SI{0.15}{m}. It is square with a side length of \SI{0.4}{m}. The target is to place the cloth on the table.
    \end{enumerate}
    The rigid object is randomly positioned for each manipulation trial in every scenario, resulting in a randomly initialized target state. The Flat Scenario is considered the simplest, while the Platform, Hemisphere, and Pole Scenarios are collectively called Complex scenarios. The Table Scenario is held out for training and serves as an unseen scenario for testing the generalizability of the proposed method.
    
    \subsection{Evaluation of the Learned Dynamics Models}
    We evaluate the effectiveness of the learned dynamics model by investigating two main aspects: whether environmental awareness in the state representation helps predict the reference trajectories and whether training the dynamics model across the four scenarios (i.e., Flat, Platform, Hemisphere, and Pole) is beneficial.
        
    \noindent \textbf{Data preparation. } We collect 10,000 rollout trajectories recorded at 100 Hz and use 90\% of the data for training and the remaining 10\% for testing. We apply a 5-frame sliding window to each trajectory to extract training samples, resulting in 200M training samples. During the test stage, we infer a trajectory from the initial state with recorded actions to measure how well the learned dynamics models can track the reference trajectory in the test set. To ensure the dynamics model applies to a wide range of fabrics and environmental setups, we randomize the physical properties of the fabric and the rigid objects in the environment. The Appendix provides more details about the simulation setup and randomization.

    \noindent \textbf{Evaluation metrics. }
    We use two metrics to measure the dynamics model's accuracy in predicting the fabric's next-time-step state: \textit{Velocity Error} and \textit{Position Error}. The former calculates the square root of the velocity error for each node in the model's graph. At the same time, the latter measures the average deviation of each node in the rollout trajectory inferred from the initial state.
    
    \vspace{-5mm}

    \begin{table}[h]
        \centering
        \resizebox{0.8\textwidth}{!}{
        \begin{tabular}{lccccc}
            \toprule
            \multirow{1}{*}{Model}& \multirow{1}{*}{Metric} &Flat &Platform&Hemisphere&Pole \\
            \midrule
        
            \multirow{2}{*}{w/o EA} & Vel. Err.~$\downarrow$ &  9.1$\pm$1.1 & 17.3$\pm$ 1.1 & 39.7$\pm$3.1 & 53.1$\pm$3.7 \\
            & Pos. Err.~$\downarrow$ & 9.7$\pm$0.8  & 19.4$\pm$1.7 & 35.3$\pm$3.5 &  46.0$\pm$3.1\\
        
            \cmidrule{1-6}
            
            \multirow{2}{*}{Specific} & Vel. Err.~$\downarrow$ &  \textbf{7.6$\pm$0.3} & 10.9$\pm$0.8& 12.3$\pm$1.1& 13.1$\pm$1.8\\
            
            & Pos. Err.~$\downarrow$ &  11.7$\pm$1.2 & 11.8$\pm$1.7& \textbf{12.6$\pm$2.1}& 12.9$\pm$2.1\\
            
            \cmidrule{1-6}
            
            
            
            \multirow{2}{*}{Pole} & Vel. Err.~$\downarrow$ &  10.6$\pm$0.8 & 19.3$\pm$1.6& 28.3$\pm$3.9& 13.1$\pm$1.8\\
            
            & Pos. Err.~$\downarrow$ &  13.5$\pm$1.2 & 28.1$\pm$2.9& 27.7$\pm$4.1& 12.9$\pm$2.1\\
            
            \cmidrule{1-6}
            
            \multirow{2}{*}{General} & Vel. Err.~$\downarrow$ & 7.9$\pm$0.6& \textbf{10.3$\pm$0.6}& \textbf{11.9$\pm$1.4}& \textbf{11.7$\pm$1.3} \\
            
            & Pos. Err.~$\downarrow$ &  \textbf{9.1$\pm$0.7}& \textbf{11.3$\pm$1.0}& 13.2$\pm$1.8& \textbf{11.5$\pm$1.4} \\
            \bottomrule
        \end{tabular}
        
        }
        \caption{Quantitative evaluation of the dynamics models shows that the dynamics model trained across different scenarios (General) consistently improves performance (unit: mm).}
        \label{tab:dynamic_evaluation}
        \vspace{-4mm}
    \end{table}

    \noindent \textbf{Result analysis. }
    Tab.~\ref{tab:dynamic_evaluation} presents the performance of the learned dynamics model in all scenarios (referred to as \textit{General}), a vanilla dynamics model without environmental awareness, and dynamics models (referred to as \textit{Specific}) specifically trained for each scenario.
        
    Firstly, the dynamics model learned without environment awareness (w/o EA) yields significantly lower results than specific and general models. This supports the choice of incorporating environmental awareness into the model design.
        
    Secondly, we compare the general dynamics model's performance with the specific models trained on their respective scenarios. The results indicate that the general model trained on a diverse set of scenarios can maintain, if not exceed, the performance of each specific model, thus validating the decision to train a general dynamics model.
        
    Finally, we apply the dynamics model trained on the Pole Scenario to the other testing scenarios. The experimental results reveal a significant domain gap between different scenarios, highlighting the need for a general model trained across various scenes to adapt the model to unseen scenarios.
    

    \begin{table}[htb]
    \centering
    \resizebox{1.0\textwidth}{!}{
    \begin{tabular}{lccc|ccc|ccc}
        \toprule
        \multirow{2}{*}{Method}  &\multicolumn{3}{c}{Flat Scenario} & \multicolumn{3}{c}{Complex Scenarios} & \multicolumn{3}{c}{Unseen Scenario}\\
        \cmidrule{2-10}
         & IoU (\%)$\uparrow$ & MPE (mm)$\downarrow$ & Time (s)$\downarrow$ 
         & IoU (\%)$\uparrow$ & MPE (mm)$\downarrow$ & Time (s)$\downarrow$ 
         & IoU (\%)$\uparrow$ & MPE (mm)$\downarrow$ & Time (s)$\downarrow$ \\
         
        \midrule
        Flingbot~\cite{ha2022flingbot}  & 85.8$\pm$13.9 & 19.3$\pm$12.2 & 2.42$\pm$ 0.13
        &49.6$\pm$32.4 & 41.4$\pm$35.3 & 2.43$\pm$ 0.12
        &51.2$\pm$23.1 & 36.2$\pm$18.4 & 2.35$\pm$ 0.15\\
    
        IRP (1st trial)~\cite{chi2022iterative}  & 80.7$\pm$3.2& 23.8$\pm$7.6 & 2.51$\pm$ 0.18
        & 40.9$\pm$13.7 & 52.4$\pm$21.6 & 2.46$\pm$ 0.14
        & 30.5$\pm$7.1 & 89.4$\pm$23.3 & 2.45$\pm$ 0.15\\
    
        IRP (2nd trial)  & 86.1$\pm$3.5& 15.3$\pm$4.9 & 4.91$\pm$ 0.28
        & 47.9$\pm$18.3 & 45.3$\pm$24.6 & 5.03$\pm$ 0.31
        & 43.2$\pm$15.9 & 40.8$\pm$15.1 & 5.01$\pm$ 0.27\\
        
        IRP (3rd Ttrial)  & \textbf{92.2$\pm$7.5}& \textbf{7.0$\pm$5.8} & 7.47$\pm$ 0.43
        & 56.4$\pm$20.4 & 39.5$\pm$27.1 & 7.25$\pm$ 0.45
        & 61.2$\pm$16.2 & 28.8$\pm$16.2 & 7.51$\pm$ 0.41\\
        \cmidrule{1-10}
        
        
        w/o EA & 90.2$\pm$7.2 & 13.4$\pm$10.2 & 2.47$\pm$ 0.15
        & 67.1$\pm$19.2 & 20.4$\pm$12.9 & 2.54$\pm$ 0.21
        & 60.4$\pm$14.1 & 29.7$\pm$18.2 & 2.39$\pm$ 0.18\\
        
        w/o MPC & 91.5$\pm$3.9 & 8.7$\pm$2.2 & \textbf{2.39$\pm$ 0.14} 
        &72.1$\pm$16.8 & 17.3$\pm$10.6 & 2.48$\pm$ 0.18
        & 72.0$\pm$15.1 & 18.2$\pm$9.3 & 2.41$\pm$ 0.17
        \\
        \cmidrule{1-10}
        
        Ours & 89.7$\pm$5.0 & 9.3$\pm$2.4 & 2.43$\pm$0.13
            & \textbf{77.2$\pm$12.8 }& \textbf{13.2$\pm$7.4} & \textbf{2.42$\pm$0.12}
        & \textbf{79.7$\pm$9.1 }& \textbf{12.4$\pm$7.9} & \textbf{2.31$\pm$0.13} \\
    
        \bottomrule
    \end{tabular}
    }

    \caption{Manipulation Evaluation and Comparison Results. IoU denotes the Intersection over Union between the actual achieved and target states. MPE denotes the Mean Pose Error that quantifies the mean particle pose error.}
    \label{tab:simulation_results}
    \vspace{-8mm}
    \end{table}
    
    \subsection{Goal-conditioned Fabric Manipulation in Simulation}
    
    After validating the dynamics model, we validate the proposed control method in goal-conditioned fabric manipulation tasks. 
        
    \noindent \textbf{Comparison with state-of-the-art methods.} We compare our method with two state-of-the-art baselines, Flingbot~\cite{ha2022flingbot} and IRP~\cite{chi2022iterative}, dedicated to fabric flinging tasks. Flingbot focuses on grasp point prediction and uses a fixed fling trajectory for all scenarios. It randomly samples 100 turning points, parameterizing 100 distinct trajectories, tested across various scenarios within the training dataset. The turning point of the best trajectory is selected as the fixed turning point and applied to the test. Conversely, IRP targets goal-conditioned fabric flinging tasks and learns delta dynamics while optimizing the action through a few trials. IRP uses a 2D image that records the movement trajectory of 9 key points of the fabric to represent the state. We report the performance of IRP with its first three trials, where the turning point of the initial trajectory for the first trial is the average point for all training data, which is consistent with the original paper.

    \noindent \textbf{Ablation study. }
    To validate the major design choices and their impact on manipulation performance, we compare our method with several ablated variants:
        
    \begin{enumerate}
        \item \textbf{w/o EA}: In this variant, we replace the environment descriptor in the node feature of our proposed method with the nodal distance and vector to the ground. As a result, this variant lacks information about its proximity to rigid objects in the three complex scenarios.
        \item \textbf{w/o MPC}: In this variant, we execute the initial trajectory from stage one without the proposed real-time control policy during the manipulation.
    \end{enumerate}
    
     \noindent \textbf{Data preparation. }
    We conducted the goal-conditioned manipulation task 20 times for each scenario to compare the abovementioned methods. The configuration of the rigid object, physical properties of the fabric, and initial and goal states are randomized in each trial within ranges specified in the Appendix.

     \noindent \textbf{Evaluation metrics. }
    The performance metrics include the Intersection over Union (IoU) between the resulting state and the target state and the Mean Pose Error (MPE), which measures the average particle pose error. Additionally, the time spent on the manipulation is reported to highlight the effectiveness of our method, which is capable of completing the task in a single attempt.

     \noindent \textbf{Result Analysis.} 
    The results of comparative and ablation studies are presented in Tab.~\ref{tab:simulation_results}. Our method demonstrates a lower Mean Pose Error (MPE) of approximately 10 mm across the simple Flat Scenario, the three Complex Scenarios, and the Unseen Scenario compared to Flingbot.

    In both Complex Scenarios and the Unseen Scenario, our method significantly outperforms IRP even after it uses 3 trials. The only exception is the Flat Scenario, where after using 3 trials, IRP achieves slightly better results than our method, albeit at the expense of longer execution time.

    Comparing our proposed method with its two variants, it becomes evident that environmental awareness and trajectory refinement through MPC are crucial for goal-conditioned tasks in complex scenarios. The absence of environmental awareness in the variant leads to a significant decrease in performance, highlighting the usefulness of environmental awareness in adapting to different or unseen environmental setups. Furthermore, the consistent improvement of our method over its variant without MPC trajectory refinement demonstrates the necessity of real-time trajectory refinement.
        
    \subsection{Real-world Experiments}
    
     \noindent \textbf{Robot setup.}
    Real-world experiments were conducted using a dual-arm setup consisting of a UR10e robot on the left and a UR16e robot on the right, equipped with 2F-85 grippers (refer to Fig.~\ref{fig:overview}). The robots were positioned 1.2 meters apart and faced each other. A single RGBD camera (Intel RealSense D435) operating at 30fps with a resolution of 640$\times$480 was used to segment the background and track the fabric. The RGB information was also utilized to efficiently segment the area of interest, including the fabric and the rigid object (e.g., the pole) in the environment. Depth information of the rigid object and fabric was calculated to the point cloud. The point cloud of a fabric piece was used to construct the graph-based state representation, as mentioned in Sec.~\ref{sec:problem statement}. The point cloud of a rigid object was used to derive $\mathbf e_i$ in the object state representation.

    \vspace{-5mm}
    
    \begin{figure}[!h]
        \centering
        \includegraphics[width=0.9\textwidth]{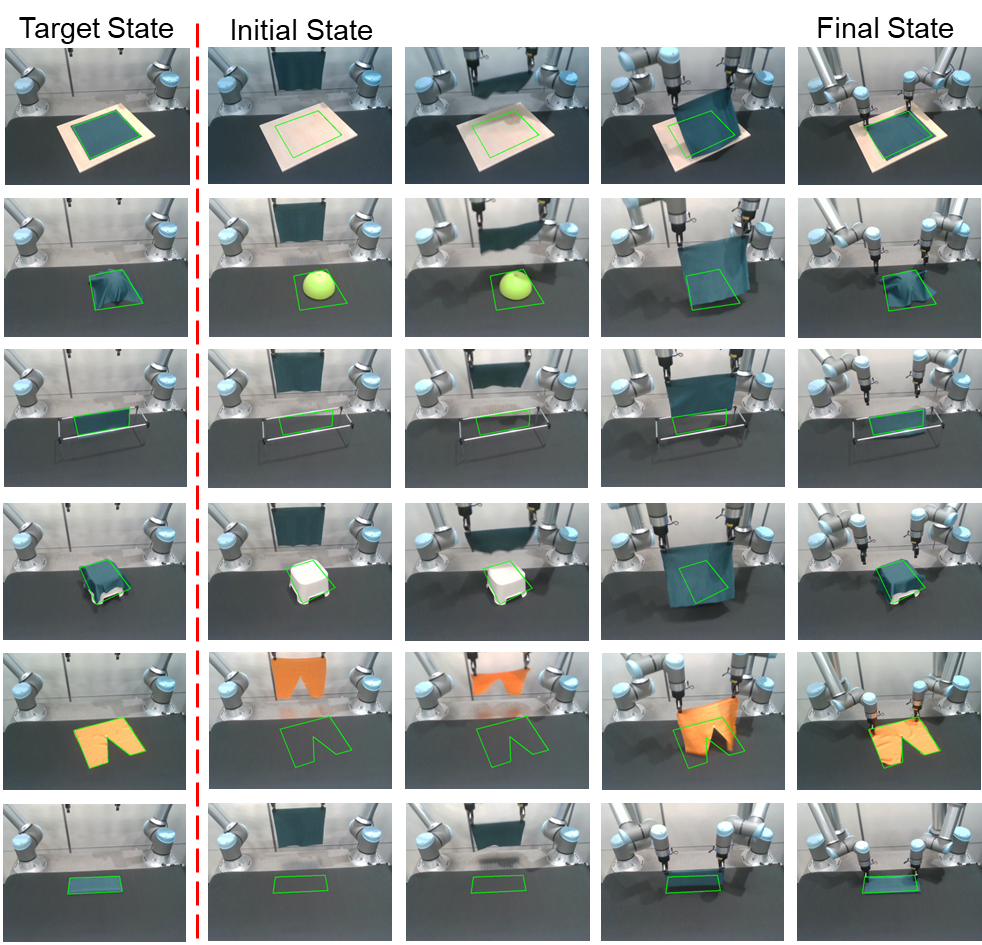}
        \caption{Qualitative results of the real-world goal-conditioned manipulation experiments. The first four rows showcase four rigid objects similar to those in the simulation, including an elevated platform, a hemisphere, a wireframe mimicking a pole, and a stool mimicking a table. The fifth row demonstrates our method's ability to generalize to an unseen fabric piece with a concave shape. In the final row, our method is applied to fold a fabric piece in half, in addition to previous unfolding or hanging tasks.
        }
        \label{fig:real-world exp}
        \vspace{-6mm}
    \end{figure}

        \begin{table}[!thb]
    \centering
    \begin{tabular}{lccccc}
        \toprule
        \multirow{2}{*}{}  &\multicolumn{4}{c}{Seen scenarios} & Unseen scenario  \\
        \cmidrule{2-6}
        
         & Flat & Platform & Hemisphere & Pole & Table/Stool \\
        \midrule
        Sim (reference)  & 9.3$\pm$2.4 & 8.5$\pm$6.1 & 10.3$\pm$2.2 & 20.7$\pm$6.1 & 12.4$\pm$7.9\\
        Real w/o MPC  & 12.3$\pm$7.9 & 13.5$\pm$5.2 & 35.1$\pm$11.6 & 33.9$\pm$17.1 & 27.1$\pm$9.1 \\
        Real  & 10.6$\pm$6.3 & 12.8$\pm$7.1 & 21.1$\pm$3.7 & 25.2$\pm$10.2 & 21.9$\pm$9.4 \\
        \bottomrule
    \end{tabular}
    \caption{Quantitative evaluation of our proposed method on real-world robot setup. 
    The error in the real-world robot experiment is measured by the double-sided Chamfer distance, while the mean particle error in the simulation is provided as a reference. (unit: mm)}
        \vspace{-6mm}
    \label{tab:real_robot_results}
    \end{table}
    
     \noindent \textbf{Acquisition of target state. }
    Before each trial, the experimenter arbitrarily positioned the rigid object and fabric. The point cloud of the fabric was then set as the target state for the subsequent manipulation. It should be noted that the target state in the real world may be partially observable due to self-occlusion. Only the polygonal contour connecting the corners of the fabric pieces was displayed for visualization purposes.
    
    In the real-world experiments, we exclusively utilized the general dynamics model for the goal-conditioned manipulation task. Commonly encountered objects with shapes similar to the environmental obstacles in our simulation data were used for training. Performance was evaluated using the Mean Pose Error (MPE), as defined in the simulation studies, across four familiar scenarios and one unseen scenario.

     \noindent \textbf{Result analysis. }
    Tab.~\ref{tab:real_robot_results} presents the performance of our method with and without real-time trajectory control in real-world settings. The performance in the real-world experiments was measured by calculating the double-sided Chamfer distance between the target point cloud and the point cloud of the fabric after execution.    
    Our method demonstrated satisfactory performance across the five scenarios in real settings. Although the Chamfer distance in real-world scenarios was generally higher than the corresponding MPE in simulations, this was expected due to the complexities and uncertainties inherent in real-world scenarios. When applied to an unseen scenario (the table), the proposed method yielded a double-sided Chamfer distance of $21.9\pm9.4$\SI{}{mm}, demonstrating its ability to generalize to new situations.    
    
    To further validate the method's generalizability to fabrics with varied physical properties, we performed the goal-conditioned manipulation task on an unseen fabric piece with a concave shape. This resulted in a double-sided Chamfer distance of $13.1\pm6.3$\SI{}{mm}. The results are shown in the fifth row of Fig.~\ref{fig:real-world exp}. It is worth noting that this fabric piece is softer and different from the other fabrics, highlighting the effective application of our method to fabrics with varying properties.
    
    In addition to the unfolding or hanging tasks presented earlier, we demonstrated that our dynamics model can be applied to a folding-in-half task because it does not rely on a specific target during training. For this task, we used the folded target state shown in the bottom row of Fig.~\ref{fig:real-world exp}. Due to severe self-occlusion, this task is challenging. However, our method achieved a double-sided Chamfer distance of $21.6\pm6.3$\SI{}{mm}.
    
    These experimental results demonstrate the proposed method's effectiveness and practical utility in real-world settings. For videos of real-world robotic fabric manipulation, please refer to supplementary materials.
    
\section{Conclusion}
\label{sec:conclusion}

    In conclusion, our research showcases a substantial advancement in robotic dynamic manipulation of fabrics within complex environments. The proposed method, incorporating environment-aware dynamics and efficient goal-conditioned manipulation, has demonstrated effectiveness in both simulated and real-world settings. The experimental outcomes underscore the method's practical application potential.
        
    However, the method has several limitations. Firstly, as we collect data from simulations, the model's accuracy is contingent upon the simulation's fidelity. This issue could be addressed with a more advanced simulator, such as~\cite{li2022diffcloth}. Secondly, the partial observation of the fabric cloth may impact the model's predictions. The point cloud quality might decrease, especially during rapid movements. Cloth reconstruction methods like~\cite{wang2023trtm} for unobserved fabric or reactive perception techniques like~\cite{zhou2024reactive, yang2021learning} could be potential solutions and areas for future research.
    
\subsubsection{\ackname} This research is partially supported by the Innovation and Technology Commission of the HKSAR Government under the InnoHK initiative, the National Natural Science Foundation of China [62206119], Shenzhen Long-Term Support for Higher Education at SUSTech [20231115141649002], and SUSTech Virtual Teaching Lab for Machine Intelligence Design and Learning [Y01331838].

%

\bibliographystyle{splncs04}
\bibliography{References}

\begin{thebibliography}{10}
\providecommand{\url}[1]{\texttt{#1}}
\providecommand{\urlprefix}{URL }
\providecommand{\doi}[1]{https://doi.org/#1}

\bibitem{berenson2013manipulation}
Berenson, D.: \href{https://ieeexplore.ieee.org/abstract/document/6697007}{Manipulation of deformable objects without modeling and simulating deformation}. In: IEEE/RSJ International Conference on Intelligent Robots and Systems (IROS). pp. 4525--4532. IEEE (2013)

\bibitem{canberk2023cloth}
Canberk, A., Chi, C., Ha, H., Burchfiel, B., Cousineau, E., Feng, S., Song, S.: \href{https://doi.org/10.1109/ICRA48891.2023.10161546}{Cloth funnels: Canonicalized-alignment for multi-purpose garment manipulation}. In: IEEE International Conference on Robotics and Automation (ICRA). pp. 5872--5879. IEEE (2023)

\bibitem{chen2022efficiently}
Chen, L.Y., Huang, H., Novoseller, E., Seita, D., Ichnowski, J., Laskey, M., Cheng, R., Kollar, T., Goldberg, K.: \href{https://doi.org/10.1007/978-3-031-25555-7_4}{Efficiently learning single-arm fling motions to smooth garments}. In: The International Symposium of Robotics Research. pp. 36--51. Springer (2022)

\bibitem{chi2022iterative}
Chi, C., Burchfiel, B., Cousineau, E., Feng, S., Song, S.: \href{https://doi.org/10.1177/02783649231201201}{Iterative Residual Policy: For Goal-Conditioned Dynamic Manipulation of Deformable Objects}. The International Journal of Robotics Research  \textbf{43}(4),  389--404 (2024)

\bibitem{ghasemipour2022blocks}
Ghasemipour, S.K.S., Kataoka, S., David, B., Freeman, D., Gu, S.S., Mordatch, I.: \href{https://proceedings.mlr.press/v162/ghasemipour22a.html}{Blocks Assemble! {L}earning to Assemble with Large-Scale Structured Reinforcement Learning}. In: Chaudhuri, K., Jegelka, S., Song, L., Szepesvari, C., Niu, G., Sabato, S. (eds.) Proceedings of the 39th International Conference on Machine Learning. Proceedings of Machine Learning Research, vol.~162, pp. 7435--7469. PMLR (17--23 Jul 2022)

\bibitem{ha2022flingbot}
Ha, H., Song, S.: \href{https://proceedings.mlr.press/v164/ha22a.html}{FlingBot: The Unreasonable Effectiveness of Dynamic Manipulation for Cloth Unfolding}. In: Faust, A., Hsu, D., Neumann, G. (eds.) Proceedings of the 5th Conference on Robot Learning. Proceedings of Machine Learning Research, vol.~164, pp. 24--33. PMLR (08--11 Nov 2022)

\bibitem{hoque2020visuospatial}
Hoque, R., Seita, D., Balakrishna, A., Ganapathi, A., Tanwani, A., Jamali, N., Yamane, K., Iba, S., Goldberg, K.: \href{https://doi.org/10.15607/RSS.2020.XVI.034}{VisuoSpatial Foresight for Multi-Step, Multi-Task Fabric Manipulation}. In: Proceedings of Robotics: Science and Systems. Corvalis, Oregon, USA (July 2020)

\bibitem{li2022diffcloth}
Li, Y., Du, T., Wu, K., Xu, J., Matusik, W.: \href{https://doi.org/10.1145/3527660}{DiffCloth: Differentiable Cloth Simulation with Dry Frictional Contact}. ACM Transactions on Graphics  \textbf{42}(1) (oct 2022)

\bibitem{lin2022learning}
Lin, X., Wang, Y., Huang, Z., Held, D.: \href{https://proceedings.mlr.press/v164/lin22a.html}{Learning Visible Connectivity Dynamics for Cloth Smoothing}. In: Faust, A., Hsu, D., Neumann, G. (eds.) Proceedings of the 5th Conference on Robot Learning. Proceedings of Machine Learning Research, vol.~164, pp. 256--266. PMLR (08--11 Nov 2022)

\bibitem{lin2021softgym}
Lin, X., Wang, Y., Olkin, J., Held, D.: \href{https://proceedings.mlr.press/v155/lin21a.html}{SoftGym: Benchmarking Deep Reinforcement Learning for Deformable Object Manipulation}. In: Kober, J., Ramos, F., Tomlin, C. (eds.) Proceedings of the 2020 Conference on Robot Learning. Proceedings of Machine Learning Research, vol.~155, pp. 432--448. PMLR (16--18 Nov 2021)

\bibitem{mason1993dynamic}
Mason, M.T., Lynch, K.M.: \href{https://ieeexplore.ieee.org/document/583093}{Dynamic manipulation}. In: Proceedings of 1993 IEEE/RSJ International Conference on Intelligent Robots and Systems (IROS'93). vol.~1, pp. 152--159. IEEE (1993)

\bibitem{mcconachie2020manipulating}
McConachie, D., Dobson, A., Ruan, M., Berenson, D.: \href{https://journals.sagepub.com/doi/full/10.1177/0278364920918299}{Manipulating deformable objects by interleaving prediction, planning, and control}. The International Journal of Robotics Research  \textbf{39}(8),  957--982 (2020)

\bibitem{niu2023goats}
Niu, Y., Jin, S., Zhang, Z., Zhu, J., Zhao, D., Zhang, L.: \href{https://ieeexplore.ieee.org/abstract/document/10342221}{Goats: Goal sampling adaptation for scooping with curriculum reinforcement learning}. In: 2023 IEEE/RSJ International Conference on Intelligent Robots and Systems (IROS). pp. 1023--1030. IEEE (2023)

\bibitem{seita2021learning}
Seita, D., Florence, P., Tompson, J., Coumans, E., Sindhwani, V., Goldberg, K., Zeng, A.: \href{https://arxiv.org/abs/2012.03385}{Learning to rearrange deformable cables, fabrics, and bags with goal-conditioned transporter networks}. In: 2021 IEEE International Conference on Robotics and Automation (ICRA). pp. 4568--4575. IEEE (2021)

\bibitem{shen2023action}
Shen, B., Jiang, Z., Choy, C., Savarese, S., Guibas, L.J., Anandkumar, A., Zhu, Y.: \href{https://doi.org/10.1177/02783649231191222}{Action-conditional implicit visual dynamics for deformable object manipulation}. The International Journal of Robotics Research  \textbf{0}(0),  02783649231191222 (0)

\bibitem{sun2023bridging}
Sun, H., Yang, L., Gu, Y., Pan, J., Wan, F., Song, C.: \href{https://www.mdpi.com/2313-7673/8/4/364}{Bridging locomotion and manipulation using reconfigurable robotic limbs via reinforcement learning}. Biomimetics  \textbf{8}(4), ~364 (2023)

\bibitem{sundaresan2020learning}
Sundaresan, P., Grannen, J., Thananjeyan, B., Balakrishna, A., Laskey, M., Stone, K., Gonzalez, J.E., Goldberg, K.: \href{https://doi.org/10.1109/ICRA40945.2020.9197121}{Learning Rope Manipulation Policies Using Dense Object Descriptors Trained on Synthetic Depth Data}. In: IEEE International Conference on Robotics and Automation (ICRA). pp. 9411--9418 (2020)

\bibitem{wang2020swingbot}
Wang, C., Wang, S., Romero, B., Veiga, F., Adelson, E.: \href{https://doi.org/10.48550/arXiv.2101.11812}{Swingbot: Learning physical features from in-hand tactile exploration for dynamic swing-up manipulation}. In: 2020 IEEE/RSJ International Conference on Intelligent Robots and Systems (IROS). pp. 5633--5640. IEEE (2020)

\bibitem{wang2018nervenet}
Wang, T., Liao, R., Ba, J., Fidler, S.: \href{https://openreview.net/forum?id=S1sqHMZCb}{Nervenet: Learning structured policy with graph neural networks}. In: International conference on learning representations (2018)

\bibitem{wang2023trtm}
Wang, W., Li, G., Zamora, M., Coros, S.: \href{https://arxiv.org/abs/2308.04670}{TRTM: Template-based Reconstruction and Target-oriented Manipulation of Crumpled Cloths}. arXiv:2308.04670 [cs.RO]  (2023)

\bibitem{weng2022fabricflownet}
Weng, T., Bajracharya, S.M., Wang, Y., Agrawal, K., Held, D.: \href{https://proceedings.mlr.press/v164/weng22a.html}{FabricFlowNet: Bimanual Cloth Manipulation with a Flow-based Policy}. In: Faust, A., Hsu, D., Neumann, G. (eds.) Proceedings of the 5th Conference on Robot Learning. Proceedings of Machine Learning Research, vol.~164, pp. 192--202. PMLR (08--11 Nov 2022)

\bibitem{xu2022dextairity}
Xu, Z., Chi, C., Burchfiel, B., Cousineau, E., Feng, S., Song, S.: \href{https://doi.org/10.48550/arXiv.2203.01197}{Dextairity: Deformable manipulation can be a breeze}. arXiv preprint arXiv:2203.01197  (2022)

\bibitem{yan2021learning}
Yan, W., Vangipuram, A., Abbeel, P., Pinto, L.: \href{https://proceedings.mlr.press/v155/yan21a.html}{Learning Predictive Representations for Deformable Objects Using Contrastive Estimation}. In: Kober, J., Ramos, F., Tomlin, C. (eds.) Proceedings of the 2020 Conference on Robot Learning. Proceedings of Machine Learning Research, vol.~155, pp. 564--574. PMLR (16--18 Nov 2021)

\bibitem{yang2021learning}
Yang, L., Han, X., Guo, W., Wan, F., Pan, J., Song, C.: \href{https://ieeexplore.ieee.org/abstract/document/9376610}{Learning-based optoelectronically innervated tactile finger for rigid-soft interactive grasping}. IEEE Robotics and Automation Letters  \textbf{6}(2),  3817--3824 (2021)

\bibitem{yang2023tacgnn}
Yang, L., Huang, B., Li, Q., Tsai, Y.Y., Lee, W.W., Song, C., Pan, J.: \href{https://doi.org/10.1109/LRA.2023.3264759}{TacGNN: Learning Tactile-Based In-Hand Manipulation With a Blind Robot Using Hierarchical Graph Neural Network}. IEEE Robotics and Automation Letters  \textbf{8}(6),  3605--3612 (2023)

\bibitem{yin2021modeling}
Yin, H., Varava, A., Kragic, D.: \href{https://www.science.org/doi/10.1126/scirobotics.abd8803}{Modeling, learning, perception, and control methods for deformable object manipulation}. Science Robotics  \textbf{6}(54),  eabd8803 (2021)

\bibitem{zeng2021transporter}
Zeng, A., Florence, P., Tompson, J., Welker, S., Chien, J., Attarian, M., Armstrong, T., Krasin, I., Duong, D., Sindhwani, V., Lee, J.: \href{https://proceedings.mlr.press/v155/zeng21a.html}{Transporter Networks: Rearranging the Visual World for Robotic Manipulation}. In: Kober, J., Ramos, F., Tomlin, C. (eds.) Proceedings of the 2020 Conference on Robot Learning. Proceedings of Machine Learning Research, vol.~155, pp. 726--747. PMLR (16--18 Nov 2021)

\bibitem{zeng2020tossingbot}
Zeng, A., Song, S., Lee, J., Rodriguez, A., Funkhouser, T.: \href{https://doi.org/10.1109/TRO.2020.2988642}{Tossingbot: Learning to throw arbitrary objects with residual physics}. IEEE Transactions on Robotics  \textbf{36}(4),  1307--1319 (2020)

\bibitem{zhang2022learning}
Zhang, F., Demiris, Y.: \href{https://doi.org/10.1126/scirobotics.abm6010}{Learning garment manipulation policies toward robot-assisted dressing}. Science Robotics  \textbf{7}(65),  eabm6010 (2022)

\bibitem{zhang2024cafknet}
Zhang, Z., Yang, L., Sun, C., Shang, W., Pan, J.: \href{https://arxiv.org/abs/2402.18420}{CafkNet: GNN-Empowered Forward Kinematic Modeling for Cable-Driven Parallel Robots}. arXiv preprint arXiv:2402.18420  (2024)

\bibitem{zhou2024reactive}
Zhou, P., Zheng, P., Qi, J., Li, C., Lee, H.Y., Duan, A., Lu, L., Li, Z., Hu, L., Navarro-Alarcon, D.: \href{https://www.sciencedirect.com/science/article/abs/pii/S0736584524000139}{Reactive human--robot collaborative manipulation of deformable linear objects using a new topological latent control model}. Robotics and Computer-Integrated Manufacturing  \textbf{88},  102727 (2024)

\bibitem{zhou2021lasesom}
Zhou, P., Zhu, J., Huo, S., Navarro-Alarcon, D.: \href{https://doi.org/10.1109/LRA.2021.3074872}{LaSeSOM: A Latent and Semantic Representation Framework for Soft Object Manipulation}. IEEE Robotics and Automation Letters  \textbf{6}(3),  5381--5388 (2021)

\end{thebibliography}

\clearpage
\begin{appendices}
    \section{Details of Dynamics Model}
    \label{app: dynamic modell}
    
    As mentioned in the main text, we use the graph-based network architecture from~\cite{lin2022learning} as the base model for our dynamics model ($G_{\text{dyn}}$). Each node $i$ is associated with an additional feature vector $\mathbf{p}_i$ in this model. For the implementation of the edge prediction model ($G_{\text{edge}}$), we follow the approach proposed in~\cite{lin2022learning}.
    
    \subsection{Input}
    The dynamics model takes the state representation of the fabric $S_t$ as input and presents it as a graph. In this graph, the nodes correspond to the points in the voxelized point cloud of the fabric captured by the depth camera. This approach helps to mitigate the sim2real gap. Each node $i$ feature concatenates its past $n=5$ velocities, a one-hot encoding of the point type (picked or not picked), and environmental information, as shown in Section~\ref{sec:problem statement}. For each edge $e_{jk}$ connecting nodes $j$ and $k$, the edge feature includes the distance vector $(\mathbf{x}_j - \mathbf{x}_k)$, its L2 norm $||\mathbf{x}_j - \mathbf{x}_k||_2$, and a one-hot encoding of the edge type (geometry edge or neighboring edge).
    
    We modify the input graph when the gripper holds the cloth to incorporate the picker action into the grasped node. We denote the picked point as $u$ and assume it is rigidly attached to the gripper. Thus, when considering the effect of the robot gripper's movement, we directly set the picked point $u$'s position $\mathbf{x}_{u,t+1} = \mathbf{x}_{u,t} + \mathbf{a}$ and velocity $\dot{\mathbf{x}}_{u,t+1} = \mathbf{a}/\Delta t$, where $\Delta t$ is the time interval for each prediction and $a$ denotes the picker action.
        
    \subsection{Architecture}
    
    The dynamics model in the paper is a graph neural network consisting of an encoder, a processor, and a decoder. The encoder utilizes two MLPs to convert node and edge features into embeddings. The processor updates these embeddings through 10 Graph Network blocks, incorporating edge, node, and global updates with residual connections to improve feature representation. The decoder then predicts point velocity based on the final node embeddings and uses these predictions to update the node positions.
        
    \subsection{Hyperparameters}
    All hyperparameters for the dynamics model and simulation setup are listed in Tab.~\ref{tab:hyperparameters}.
    
        \begin{table}[!htb]
        \centering
        \begin{tabular}{ll}
        \hline
        \textbf{Model Parameter} & \textbf{Value} \\
        \hline
        \multicolumn{2}{c}{Encoder} \\
        \hline
        Number of hidden layers & 3 \\
        Size of hidden layers & 128 \\
        \hline
        \multicolumn{2}{c}{Processor} \\
        \hline
        Number of message passing steps & 10 \\
        Number of hidden layers in each edge/node update MLP & 3 \\
        Size of hidden layers & 128 \\
        \hline
        \multicolumn{2}{c}{Decoder} \\
        \hline
        Number of hidden layers & 3 \\
        Size of hidden layers & 128 \\
        \hline
        \multicolumn{2}{c}{Training Parameters} \\
        \hline
        Learning rate & 0.0001 \\
        Batch size & 16 \\
        Training epoch & 50 \\
        Optimizer & Adam \\
        Beta1 & 0.9 \\
        Beta2 & 0.999 \\
        Weight decay & 0 \\
        \hline
        \multicolumn{2}{c}{Others} \\
        \hline
        dt & 0.01 second \\
        Particle radius & 0.00625 m \\
        Downsample scale & 3 \\
        Voxel size & 0.0216 m \\
        Neighbor radius $R$ & 0.045 m \\
        \hline
        \end{tabular}
        \caption{Hyperparameters for Dynamics Model}
        \label{tab:hyperparameters}
    \end{table}
    
    \section{Data Collection}
    \label{app: data collection}
    
    \begin{figure}[!thb]
          \centering
          \includegraphics[width=0.9\textwidth]{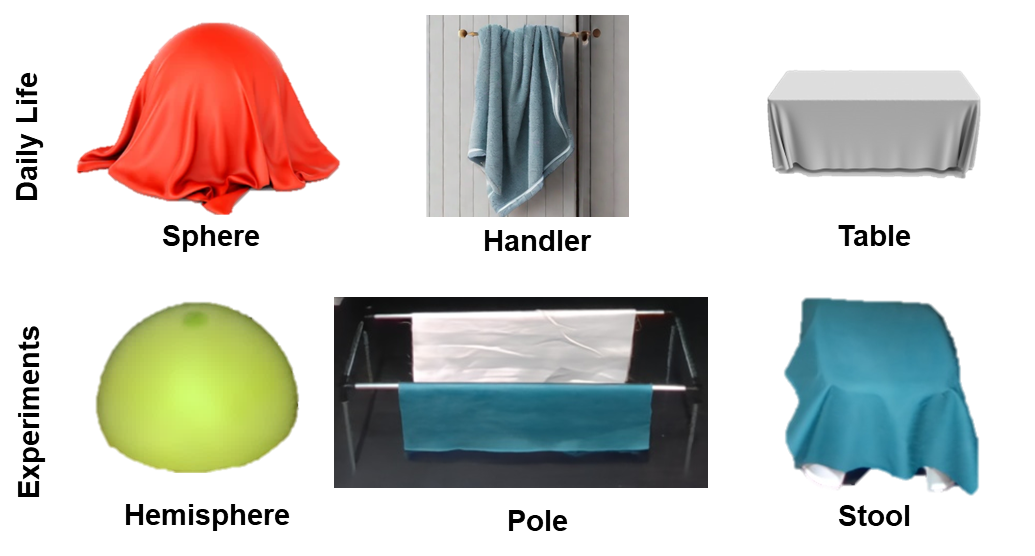}
          \caption{Cloth manipulation scenarios in daily life and experiments.}
          \label{fig:app-env}
          \vspace{-3mm}
    \end{figure}
    
    As mentioned in Section~\ref{sec:dynamics}, the data is collected in simulation. The hyperparameters of the simulation setup can be found in Tab.~\ref{tab:hyperparameters}.
        
    \subsection{Bi-Partite Graph Matching}
    We employ bi-partite graph matching to map each voxelized point cloud point $\mathbf p_i$ to a corresponding simulated cloth particle $\mathbf x_j$. This matching process establishes the correspondence between the $N$ points in the point cloud $\{\mathbf p_i\}$ and the $M$ simulated cloth particles $\{\mathbf x_j\}$. 
    
    To improve computational efficiency, we downsample the simulated cloth mesh by a factor of three. For example, if the original cloth consists of $40 \times 40$ particles, it is downsampled to $13 \times 13$ particles.
    
    The bipartite graph connects each point cloud point $\mathbf p_i$ to each simulated particle $\mathbf x_j$. The cost of each edge is determined by the distance between the connected points. Notably, distances exceeding a certain threshold are disregarded to eliminate outlier points.
        
    \subsection{Domain Randomization in Dataset}
    
    To ensure that the trained dynamics model can generalize to various types of cloth, we introduce some randomness in the cloth mass, physical properties, size, and the initial pose of rigid objects in the environment. Tab.~\ref{tab:cloth_properties} presents the randomization range for each parameter.
    
    \begin{table}[h]
        \centering      
        
        \begin{tabular}{ll}
        \toprule
        Cloth Properties & Randomization range \\
        \midrule
        Mass (kg) & $\mathcal{U}(0.05, 0.5)$ \\
        Size (m) & $\mathcal{U}(0.2, 0.4)$ \\
        Stretchability & $\mathcal{U}(0.5, 2.0)$ \\
        Bendability & $\mathcal{U}(0.5, 2.0)$ \\
        Shearability & $\mathcal{U}(0.5, 2.0)$ \\
        \midrule
        Rigid Object  & Randomization range \\
        \midrule
        z-axis coordinate (m) &   $\mathcal{U}(0.1, 0.3)$\\
        y-axis coordinate (m) &   $\mathcal{U}(-0.2, 0.2)$\\
        Rotation angle about z-axis  \ \ \    &   $\mathcal{U}(-30^\circ, 30^\circ)$\\
        \bottomrule
        \end{tabular}
        \caption{Randomization range for cloth properties and rigid objects in the environment.}
        \label{tab:cloth_properties}
        
    \end{table}
    
    \section{Implementation Details for Baselines}
    \label{app: baseline}
    
    This section details the baseline methods used for comparative analysis in our study.
        
    \subsection{Flingbot Implementation}
    Flingbot uses a simple yet effective approach to selecting flying trajectories. The process starts by randomly generating $100$ distinct flying trajectories. These trajectories undergo a series of tests across different scenarios in the training dataset.
    
    The best trajectory is selected based on the performance across the tested scenarios. This optimal trajectory is then applied to the test dataset, and the performance of the fixed trajectory is evaluated on the unseen data.
    
    \subsection{Iterative Residual Policy (IRP)}
    The Iterative Residual Policy (IRP) represents the state of a trajectory as a $256 \times 256$ image projected onto the $y-z$ plane. The pixel values correspond to the occupancy probability of nine key points on the cloth. Similarly, any environmental obstacles are projected onto the $y-z$ plane.
    
    IRP's main mechanism involves learning delta dynamics, which captures the incremental changes in the state representation. This learning is then used to optimize the action iteratively. The policy's performance is evaluated after three trials.


    \begin{figure}[!thb]
          \centering
          \includegraphics[width=\textwidth]{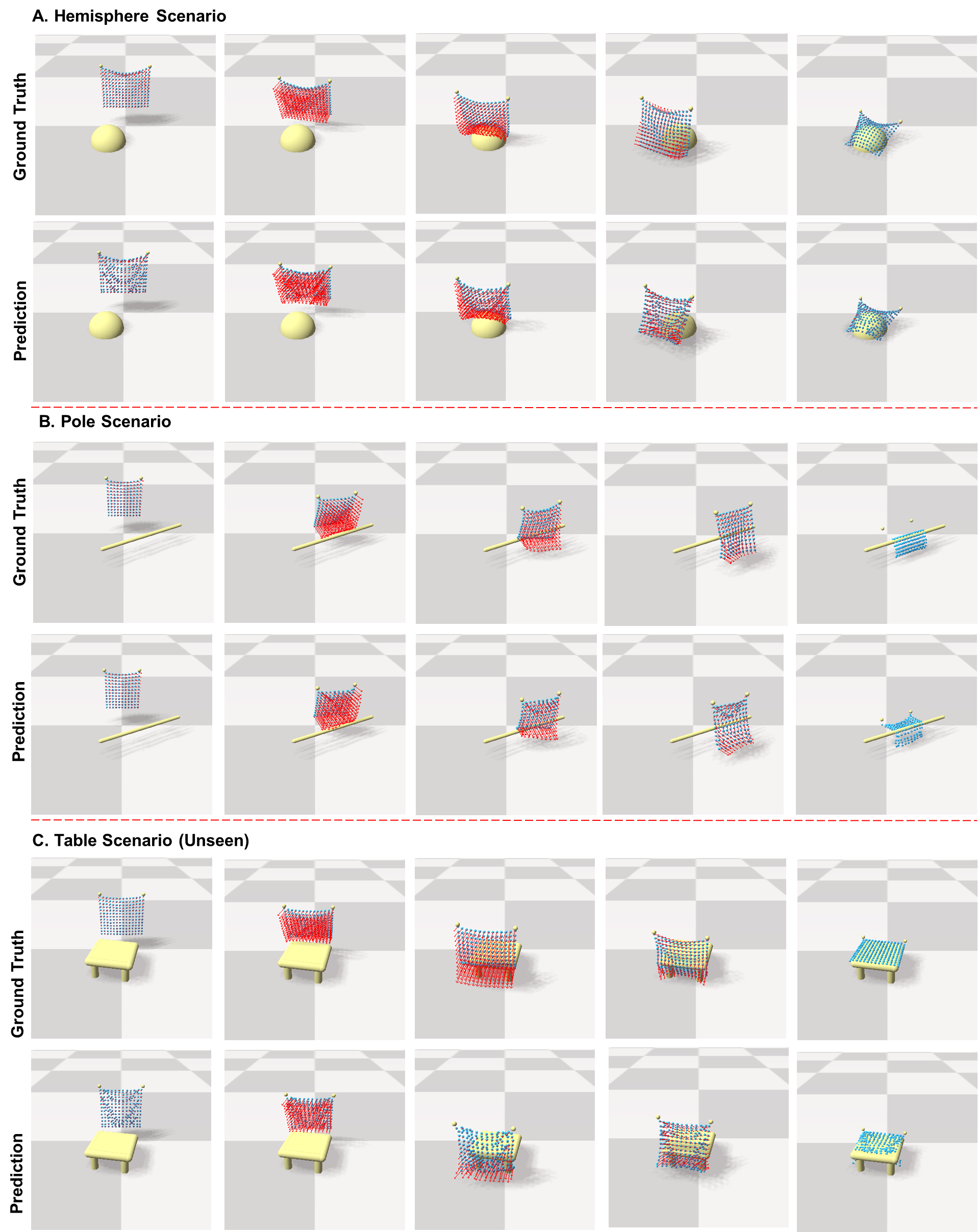}
          \caption{\textbf{Qualitative results for evaluating the learned dynamics model in three complex scenarios}: Hemisphere, Pole, and Table (unseen). Each node's predicted or ground-truth velocity is visualized as red arrows, with the arrow's length representing the velocity's magnitude.}
          \label{fig:dynamic_visual}
          \vspace{-3mm}
    \end{figure}

    \begin{figure}[!thb]
          \centering
          \includegraphics[width=\textwidth]{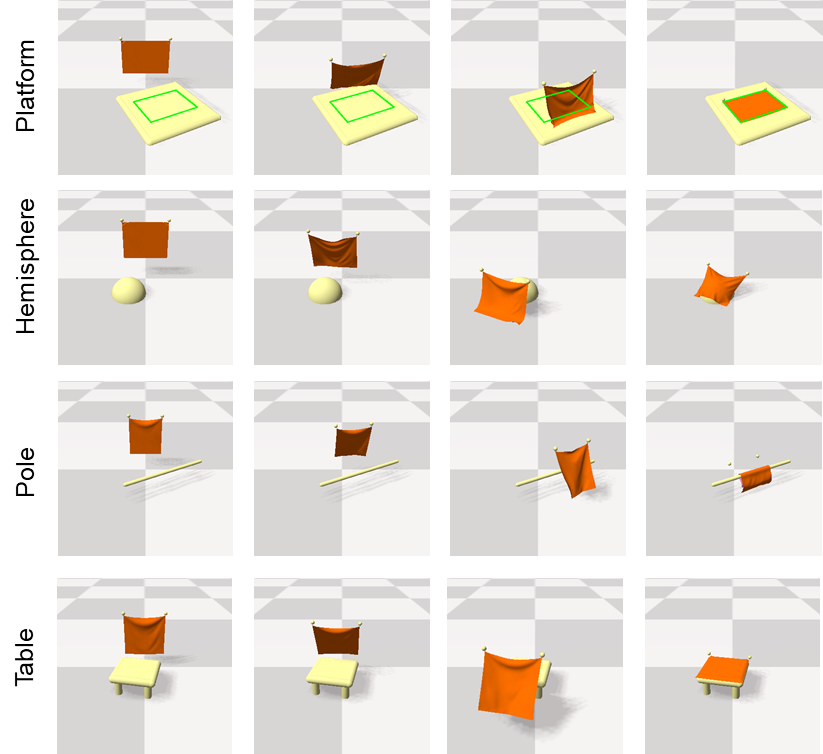}
          \caption{\textbf{Qualitative results for the goal-conditioned manipulation task.} Rollout trajectories are visualized for four complex scenarios: Platform, Hemisphere, Pole, and Table (unseen).}
          \label{fig:manipulation}
          \vspace{-5mm}
    \end{figure}
\end{appendices}
    
\end{document}